\documentclass[a4paper, 10 pt, conference]{ieeeconf} 

\IEEEoverridecommandlockouts                     

\usepackage{array}
\usepackage{url}
\usepackage{verbatim}
\usepackage{cite}
\usepackage{booktabs}
\usepackage{multirow}
\usepackage{mathtools}
\usepackage{bm} 
\usepackage{comment}
\usepackage{amsmath,amssymb,amsfonts}
\usepackage{algorithm}
\usepackage{algpseudocode}
\usepackage{graphicx}
\usepackage{textcomp}
\usepackage{hyperref}
\usepackage{stfloats}
\usepackage{xcolor}
\usepackage{paralist}
\usepackage{caption}
\usepackage{subcaption}

\newcommand{\cmmnt}[1]{}
\def\BibTeX{{\rm B\kern-.05em{\sc i\kern-.025em b}\kern-.08em
    T\kern-.1667em\lower.7ex\hbox{E}\kern-.125emX}}

\title{\LARGE \bf{Cloud-Assisted Remote Control for Aerial Robots: From Theory to Proof-of-Concept Implementation}}

\author{Achilleas Santi Seisa$^{1,*}$, Viswa Narayanan Sankaranarayanan$^{1}$, Gerasimos Damigos$^{2}$,\\Sumeet Gajanan Satpute$^{1}$ and George Nikolakopoulos$^{1}$
\thanks{This work has been partially funded by the European Union, under the Grant Agreement
No. 101139257 (SUNRISE-6G).}
\thanks{$^{1}$ The authors are with the Robotics and AI Team, Department of Computer, Electrical and Space Engineering, Lule\aa\,\, University of Technology, Lule\aa\,\,}
\thanks{$^{2}$ The author is with Ericsson Research, Lule\aa\,\,} 
\thanks{$^{*}$Corresponding Author's email: {\tt\small achsei@ltu.se}}
}

\begin{document}

\maketitle
\thispagestyle{empty}
\pagestyle{empty}

\begin{abstract}
Cloud robotics has emerged as a promising technology for robotics applications due to its advantages of offloading computationally intensive tasks, facilitating data sharing, and enhancing robot coordination. However, integrating cloud computing with robotics remains a complex challenge due to network latency, security concerns, and the need for efficient resource management. In this work, we present a scalable and intuitive framework for testing cloud and edge robotic systems. The framework consists of two main components enabled by containerized technology: (a) a containerized cloud cluster and (b) the containerized robot simulation environment. The system incorporates two endpoints of a User Datagram Protocol (UDP) tunnel, enabling bidirectional communication between the cloud cluster container and the robot simulation environment, while simulating realistic network conditions. To achieve this, we consider the use case of cloud-assisted remote control for aerial robots, while utilizing Linux-based traffic control to introduce artificial delay and jitter, replicating variable network conditions encountered in practical cloud-robot deployments.
\end{abstract}

\begin{keywords}
Robotics; Cloud Computing; Cloud Robotics.
\end{keywords}

\section{Introduction}
\label{sec:intro}
Robots are becoming increasingly integrated into various domains, from industrial automation and healthcare to autonomous vehicles and smart infrastructure~\cite{nikolakopoulos2021pushing}. However, their ability to perform complex tasks autonomously is often limited by onboard computational resources, power constraints, and real-time processing requirements~\cite{robotics7030047}. To address these challenges, cloud, and edge computing have emerged as promising solutions by enabling offloading of computationally intensive tasks, facilitating data sharing, and enhancing robot coordination~\cite{hu2012cloud}. Despite these advantages, integrating cloud computing with robotics remains a complex challenge due to network latency, security concerns, and the need for efficient resource management. Therefore, ongoing research aims to address these challenges.

Several studies have explored cloud robotics and edge robotics applications, primarily involving the Robot Operating System (ROS). For instance,~\cite{luo2017design, miratabzadeh2016cloud, rahimi2017industrial} seek to benefit from these technologies and enhance the robotic applications. Similarly,~\cite{rosa2014towards} presents a system enabling a ground robot to autonomously navigate data center rooms and collect useful measurements.

Notable implementations include offloading tasks such as simultaneous localization and mapping (SLAM)~\cite{chen2021fogros, huang2022edge, okumucs2018exploring, dechouniotis2022edge} and enabling cloud-assisted robotic missions~\cite{mehrooz2019system, wang2021mobility}. Similarly, in~\cite{wang2021mobility}, a swarm of Unmanned Aerial Vehicles (UAV)s offloads computation tasks to a mobile edge server using a deep reinforcement learning model while ensuring latency constraints are met. Meanwhile,~\cite{sun2022resource} and~\cite{wang2022network} propose deploying a powerful UAV as an assisted edge computing node.

\begin{figure}[t]
    \centering
    \includegraphics[width=\linewidth]{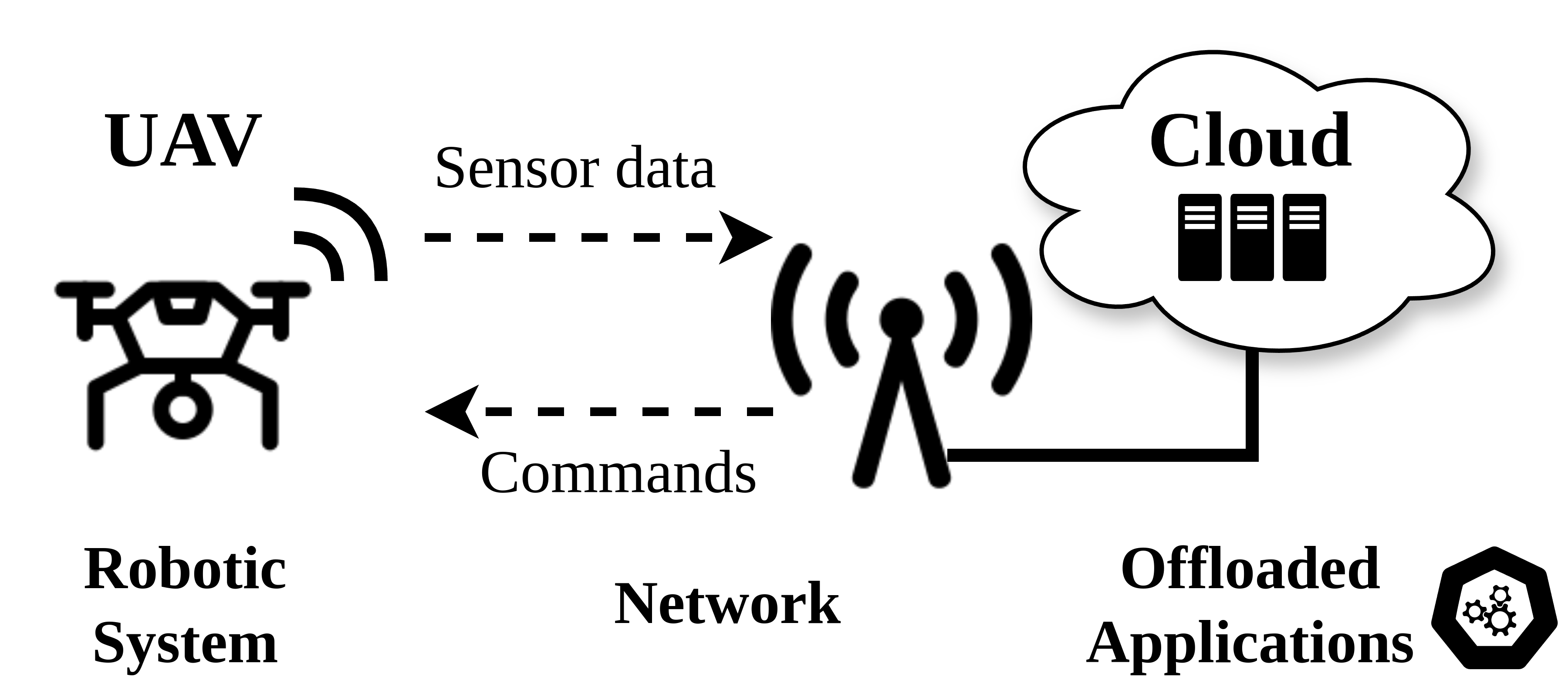}
     \caption{Illustration of a cloud robotics system.}
     \label{fig:concept}
\end{figure}

The rapid advancement of virtualization and containerized application technologies has significantly assisted cloud and edge robotics. Containerized applications must be designed to minimize edge latency and optimize resource provisioning, as discussed in~\cite{seisa2022edge2, hu2022cec}. Effective orchestration is also crucial, leading to the use of Kubernetes~\cite{seisa2022kubernetes}. In~\cite{xiong2018extend}, the authors suggest an infrastructure to connect the cloud to the edge, extending cloud services and Kubernetes management through a dedicated network protocol. Similarly,~\cite{8651759} implements a remote controller as a Docker containerized application running on a mobile edge server. Offloaded control on cloud or edge servers has also been proposed in~\cite{aarzen2018control, skarin2020cloud2}, showcasing the advantages of these technologies.

Further,~\cite{kochovski2019architecture} introduces an automated container distribution process across cloud, edge, and fog environments using Kubernetes and stochastic modeling. A framework combining Docker, Kubernetes, and ROS for monitoring containerized robot tasks is proposed in~\cite{9474167}. Likewise, a comparison between the usage of standalone Docker edge deployment, and the integration of Docker with Kubernetes for the trajectory control of an aerial robot is presented in~\cite{seisa2022comparison}. Other works such as~\cite{9568376} presents mechanisms for deploying robotic containerized applications to the edge and cloud. Lastly,~\cite{10160632} introduces KubeROS, a comprehensive framework designed to address the challenges of deploying complex robotic software in large-scale systems.

While these works demonstrate the potential of cloud robotics, existing solutions often focus on specific applications without providing a generalized, scalable framework for seamless cloud-robot collaboration. Furthermore, there is a lack of standardized, easy-to-deploy testbed that allow researchers to experiment with cloud-assisted robotic control consistently.

Identifying the growing significance of integrating cloud computing with robotics, this work presents a framework as illustrated in Fig.~\ref{fig:concept}, and establishes a baseline for cloud robotics and offers the following \textbf{contributions}:
\begin{inparaenum}
    \item a fundamental understanding of cloud robotics;
    \item an intuitive framework to emulate the cloud-robot online collaboration;
    \item a testbend for simulating a cloud-assisted remote control mission for aerial robots.
\end{inparaenum}

\section{Cloud Emulated Framework}
\label{sec:framework}
This work proposes a cloud-emulated framework for cloud robotics. The framework consists of two main components enabled by containerized technology: (a) a containerized cloud cluster and (b) a containerized robot simulation environment. A network emulation module is used, based on Linux traffic control, to simulate real-world network conditions.

\begin{figure*}[http]
    \centering
    \includegraphics[width=\linewidth]{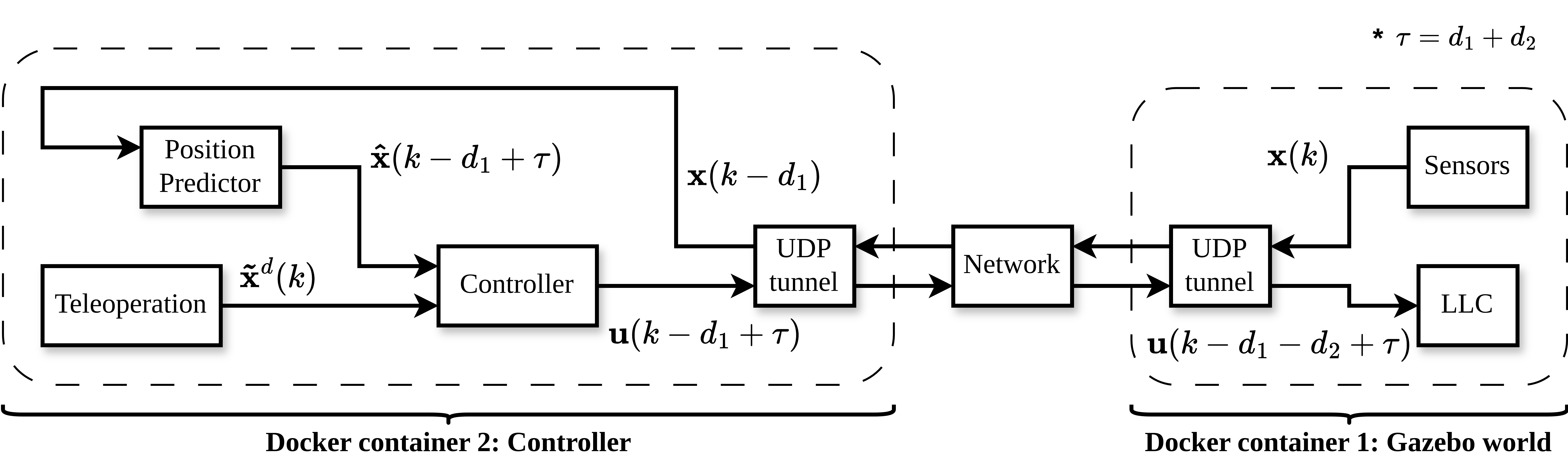}
     \caption{Block diagram of the proposed framework. It consists of an emulated cloud cluster (container 2) hosting offloaded applications, a simulated world representing the robotic environment (container 1), and a network connecting the two.}
     \label{fig:block_diagram}
\end{figure*}

\subsection{Cloud Emulation}
\label{sec:cloud}
To emulate a cloud or edge robotics environment, the framework requires three key elements~\cite{seisa2022edge}: a cloud cluster, a robotic system, and a communication channel connecting them, as shown in Fig.~\ref{fig:block_diagram}. To ensure modularity, we deploy these components as separate containers.

The first container simulates the robotic environment, including the aerial robot along with its Low Level Controller (LLC), its sensors and an end of the User Datagram Protocol (UDP) tunnel and its surroundings, while the second container emulates the cloud cluster, hosting applications responsible for the remote control of the aerial robot. These applications include the position predictor, the teleoperation module, the controller, and the other end of the UDP tunnel, as will be discussed further. These containers operate independently, allowing for flexible deployment and scalability.

\subsection{World Simulation}
\label{sec:simulation}
\begin{figure*}[http]
    \centering
    \includegraphics[width=\linewidth]{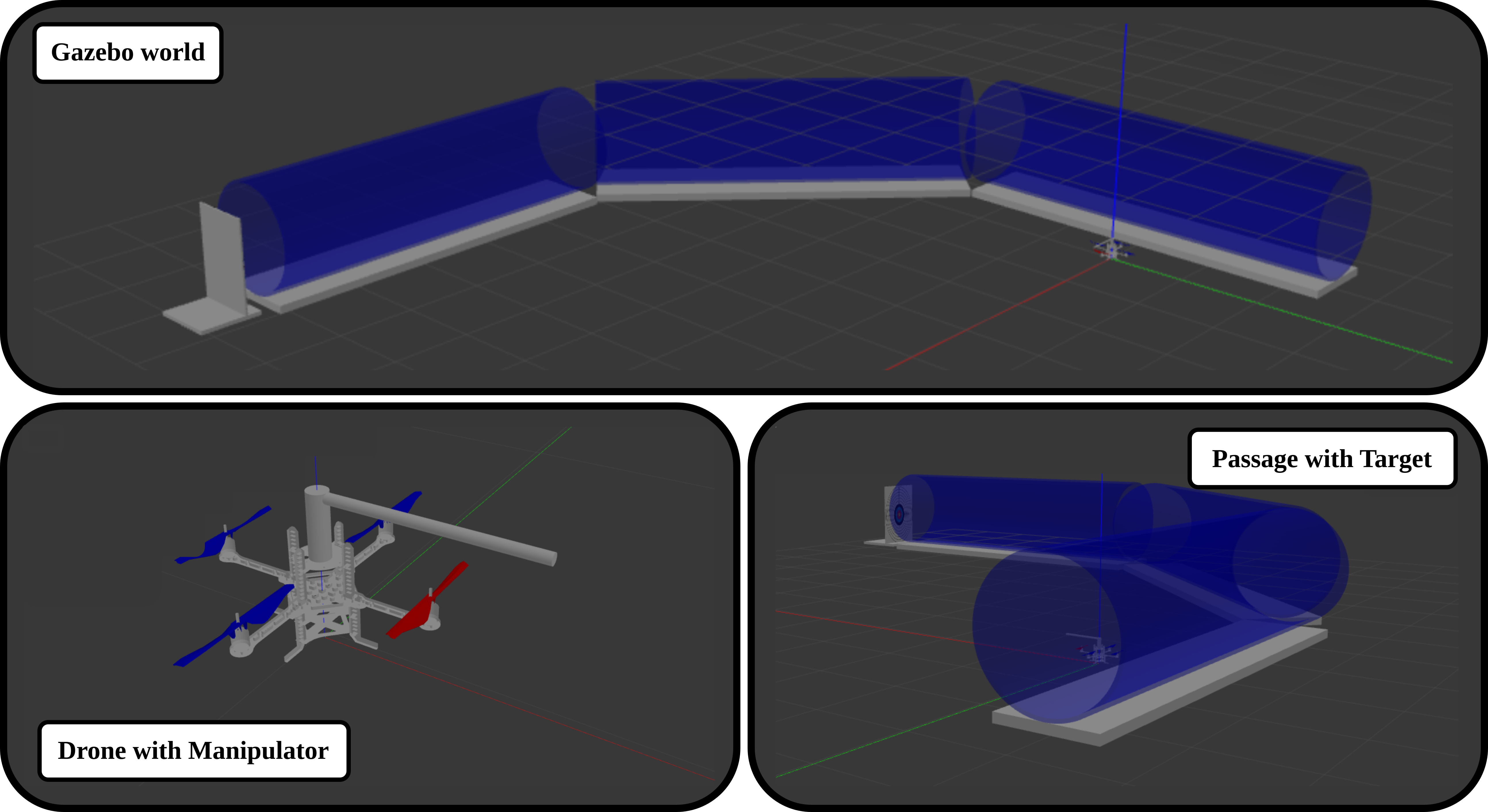}
     \caption{Illustration of a world simulation environment.}
     \label{fig:gazebo}
\end{figure*}

The simulation container (container 1) incorporates the aerial robot and its surrounding environment, as depicted in Fig.~\ref{fig:gazebo}. It hosts the Gazebo simulator, which is a simulation tool to represent physical robotic systems virtually, with a pre-defined environment, designed to evaluate cloud-assisted aerial manipulation. The environment includes: (a) a drone model from the RotorS simulator package~\cite{Furrer2016} along with its sensors and LLC, (b) a rigid manipulator (robotic arm for physical interactions) mounted on the drone for interaction tasks, (c) a narrow tube passage that challenges the drone’s navigation capabilities, and (d) a target object located at the end of the passage, requiring the drone to interact with it.

Additionally, the container simulates the drone's sensors to generate the current state $\mathbf{x}(k)$ of the aerial robot, and the LLC, that receives the control action $\mathbf{u}(k-d_1-d_2+\tau)$ from the cloud cluster and translate it into motor speed. Further, it incorporates one endpoint of a UDP tunnel, enabling bidirectional communication with the cloud cluster container while simulating realistic network conditions. The code for the simulated world and the container that hosts it can be found in the \href{https://github.com/achilleas2942/summer_school_world}{World} GitHub repository.

\subsection{Remote Control for Aerial Manipulation}
\label{sec:control}
The cloud cluster container incorporates the velocity controller for the drone, which generates the control action $\mathbf{u}(k-d_1+\tau)$. It also includes the teleoperation module to allow human input, denoted as $\mathbf{\tilde{x}}^d(k)$. In addition, it runs a position predictor to estimate the drone's current, and future states $\mathbf{\hat{x}}(k-d_1+\tau)$, based on the delayed state $\mathbf{x}(k-d_1)$, and the closed-loop delay $\hat{\tau}$. Finally, it contains one end of a UDP tunnel, which facilitates real-time communication with the simulation container while allowing us to introduce and simulate communication delays. The code for the containerized cloud cluster can be found in the \href{https://github.com/achilleas2942/summer_school_controller}{Controller} GitHub repository.

Unlike previous works such as~\cite{sankaranarayanan2023paced}, where a model predictive control methodology to compensate for time delays were developed, in this work we focus on presenting a scalable and intuitive framework for testing cloud and edge robotic systems. Therefore, for this work, we use a Proportional–Integral–Derivative (PID) controller for the cloud-assisted control.

The solution for the cloud-assisted remote control problem is divided into three steps: (a) the position predictor, (b) the teleoperation, and (c) the controller, which are further elaborated in the following subsections.

\subsubsection{Position Predictor}
\label{sec:predictor}
Since the inputs to the aerial robot are delayed by time, $\tau$, where $\tau = d_1 + d_2$ a position predictor has to be designed to predict the future estimate of the state for the current observation of the states based on the available information. The architecture is designed in such a way that the onboard PC loops back the time-stamped control input to the position predictor along with the odometry of the aerial robot. The predictor finds an estimate of the closed-loop delay, $\hat{\tau}$ in the network, using the following formulation,
\begin{align}
    \hat{\tau}(k+1) &= \hat{\tau}(k) + \frac{(\tau_{n} - \hat{\tau}(k))}{k+1}, \label{eq:delay_est} \\
    \text{where } \tau(0) &= 0 \nonumber
\end{align}
$\tau_{n}$ is the difference between the current time-stamp and the time-stamp of the delayed received control action.

Let's assume the state of the drone $\mathbf{x}$ consists of its position $\mathbf{p}$, its velocity $\mathbf{\dot{p}}$, orientation $\mathbf{q}$, and angular velocity $\mathbf{\dot{q}}$. To predict the future position and velocity for the current observations, we need to satisfy~\eqref{eq:est}.
\begin{subequations}
\label{eq:est}
    \begin{align}
        \mathbf{\hat{p}}(k) &= \mathbf{p}(k-\tau), \label{eq:pos_est} \\
        \implies \mathbf{\dot{\hat{p}}}(k) &= \mathbf{\dot{p}}(k-\tau). \label{eq:vel_est}
    \end{align}
\end{subequations}
Therefore, future estimates of the position and velocity can be calculated from \eqref{eq:est2} using the relationship,
\begin{subequations}
\label{eq:est2}
    \begin{align}
        \mathbf{{\hat{p}}}(k + \tau) &= \mathbf{\hat{p}}(k) + \mathbf{\dot{p}}(k) \tau. \label{eq:pos_est_fut} \\
        \mathbf{\dot{\hat{p}}}(k + \tau) &= \mathbf{\dot{\hat{p}}}(k) + \mathbf{\ddot{p}}(k) \tau. \label{eq:vel_est_fut_2}
    \end{align}
\end{subequations}
Similarly, we apply for the orientation of the drone.

The overall outcome of the position predictor will include the corrected state estimates $\mathbf{\hat{x}}(k+\tau)$ based on the measured delays. Hence, we are able to compensate for the introduced uplink and downlink delays.

\subsubsection{Teleoperation}
\label{sec:teleoperation}
For the teleoperation module, we consider a human input $\mathbf{\tilde{x}}^d(k)$. We allow the human operator to insert the next waypoint for the drone. This input is applied to the controller to generate corresponding control actions.

\subsubsection{Controller}
\label{sec:controller}
We develop a controller for the robot to track waypoints specified by an operator, even in the presence of time-varying delays. The controller features a dual-loop design: (a) the inner loop manages the attitude dynamics, and (b) the outer loop handles the position dynamics. Because the inner loop needs to operate at a significantly higher frequency than the outer loop to achieve the required moments for the control inputs, its control execution is done onboard, minimizing processing and actuation delays. The outer loop control is executed on a cloud cluster.

To enable cloud-assisted teleoperation and autonomous control of the aerial manipulator, we implement a PID controller for waypoint tracking. The controller receives state estimates from a position predictor and teleoperation commands, generating velocity-based control actions for the aerial robot.

The velocity-based controller operates in a feedback loop, continuously adjusting the drone’s motion based on the difference between its current state and the desired setpoint. Specifically, the controller computes the position errors as described in~\eqref{eq:error}.
\begin{align}
    \mathbf{e}_p(k) &= \mathbf{\hat{p}}(k+\tau) - \mathbf{\tilde{p}}^d(k) \label{eq:error}
\end{align}
Sequentially, the error $\mathbf{e}_p(k)$ is utilized to compute the velocity commands using a PID control law with proportional $\mathbf{K}_P$, integral $\mathbf{K}_I$ and derivative $\mathbf{K}_D$ gains, as expressed in~\eqref{eq:control}.
\begin{align}
    \mathbf{\dot{p}}^d(k) &= \mathbf{K}_P\cdot\mathbf{e}_p(k) + \mathbf{K}_I\cdot\int\mathbf{e}_p(k)dt + \mathbf{K}_D \cdot\frac{d\mathbf{e}_p}{dt}(k) \label{eq:control}
\end{align}
%

\subsection{Network Emulation}
\label{sec:network}
The main challenge in cloud robotics is network latency, which can impact real-time control and decision-making~\cite{taleb2017multi}. Since the proposed framework is tested and provided for single machine operation, it is crucial to simulate real-world latency conditions. To achieve this, we utilize Linux-based traffic control to introduce artificial delay and jitter, replicating variable network conditions encountered in practical cloud-robot deployments, while communication within the cloud cluster container and the simulated world is separated. 

\subsubsection{UDP Tunnel}
\label{sec:udp_tunnel}
For transmitting ROS messages between the aerial robot environment and the cloud, a UDP tunnel with two ends is employed. This tunnel has one endpoint operating within the simulated world container and the other in the cloud cluster container. The aerial robot transmits positional data, comprising its position, velocity, and orientation, represented as $x(k)$, where $k$ indicates the current time step. Prior to transmission, ROS messages are converted into byte arrays to simplify data transfer. The UDP tunnel extracts the data from these byte arrays and conveys it as ROS messages to the nodes within the cloud cluster container. These messages experience an uplink delay represented by $d_{1}$, rendering the positional data as $x(k-d_{1})$. Likewise, control inputs, indicated as $u(k-d_{1}-d_{2})$, with $d_{2}$ representing downlink delays, are converted into byte arrays and dispatched to the aerial robot via the UDP tunnel from the cloud.

\subsubsection{Robotic Operating System}
\label{sec:ros}
Within the cloud cluster container, communication depends on the ROS networking protocol. Every application in the cluster exists within a unified network that allows ROS nodes to interact seamlessly via ROS subscribing and publishing methods.

\subsection{Resource Utilization}
\label{sec:resources}
As the proposed framework is built on containerized technologies, it is highly adaptable and platform-agnostic while supporting multiple key features. One such feature is resource allocation on demand, allowing users to specify limits and requests for each container. This capability enables the restriction of resources for the aerial robot container, simulating real-world resource-constrained robotic platforms. Conversely, resource limits and requests can be defined for the cloud container to emulate background workloads, facilitating a more realistic simulation of cloud robotic scenarios.

\section{Experimental Evaluation}
\label{sec:experiments}
To evaluate the performance and adaptability of the proposed framework, we deploy two Docker containers on a single machine: one for the simulated world and another for the cloud cluster. Communication between these containers is established through a UDP tunnel, ensuring a bidirectional exchange of data. The whole framework is provided through the \href{https://github.com/AERO-TRAIN/exercises_summer_school_hri_day}{Cloud-enabled Remote Control} GitHub repository.

\begin{figure}[http]
    \centering
    \includegraphics[width=\linewidth]{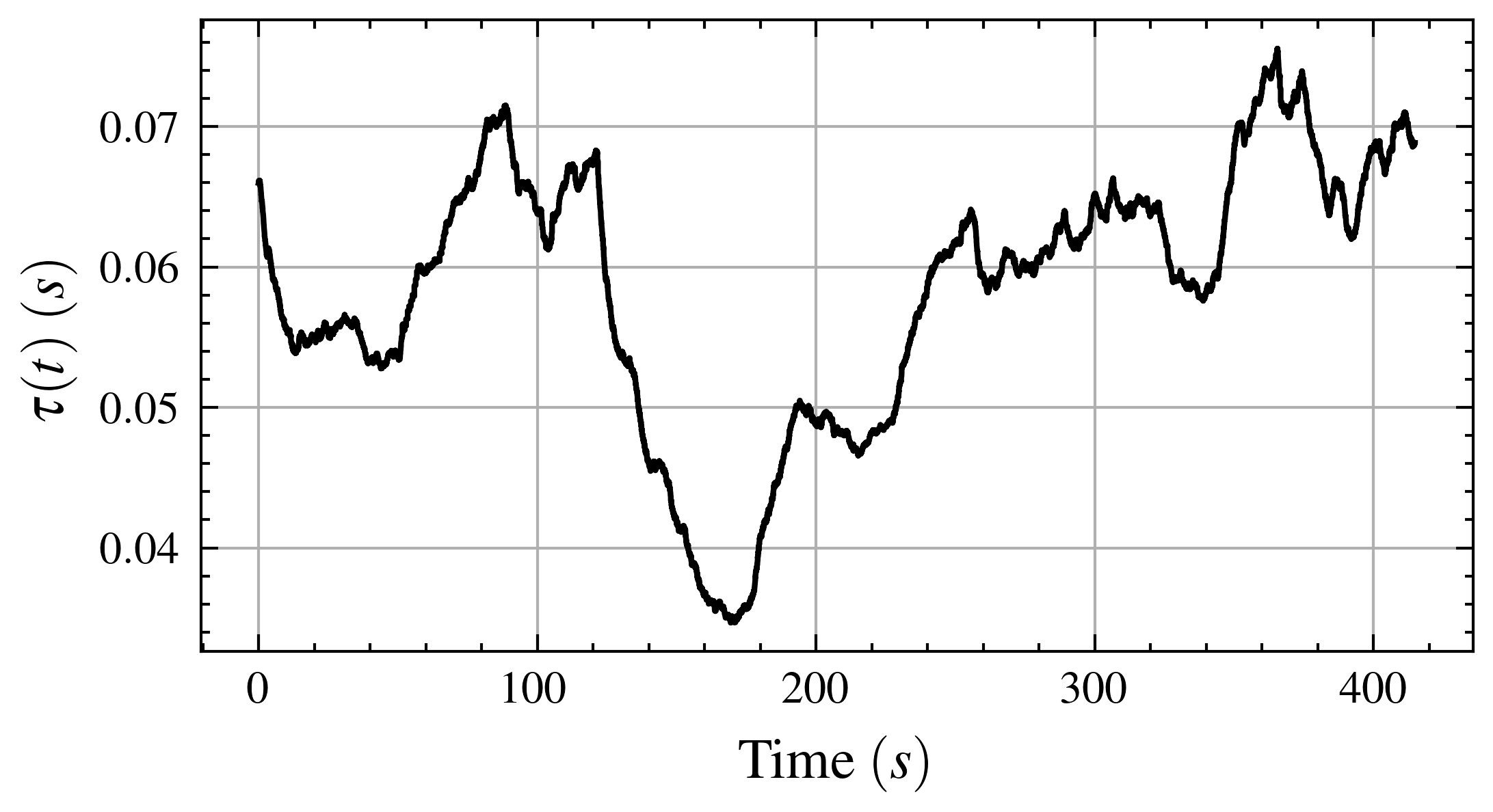}
     \caption{Round-trip time delay throughout the mission.}
     \label{fig:delay}
\end{figure}

To simulate realistic network conditions, we introduce varying latency and jitter using Linux-based traffic control (tc). We configure multiple network profiles with delays ranging from 20 to 70 ms and jitter values between 10 and 40 ms, as depicted in Fig.\ref{fig:delay}. The applied delay profile utilizes a sliding average filter, following Eq.\eqref{eq:delay_est}, to mitigate transient fluctuations and filter out peaks.

\begin{figure}[http]
    \centering
    \includegraphics[width=\linewidth]{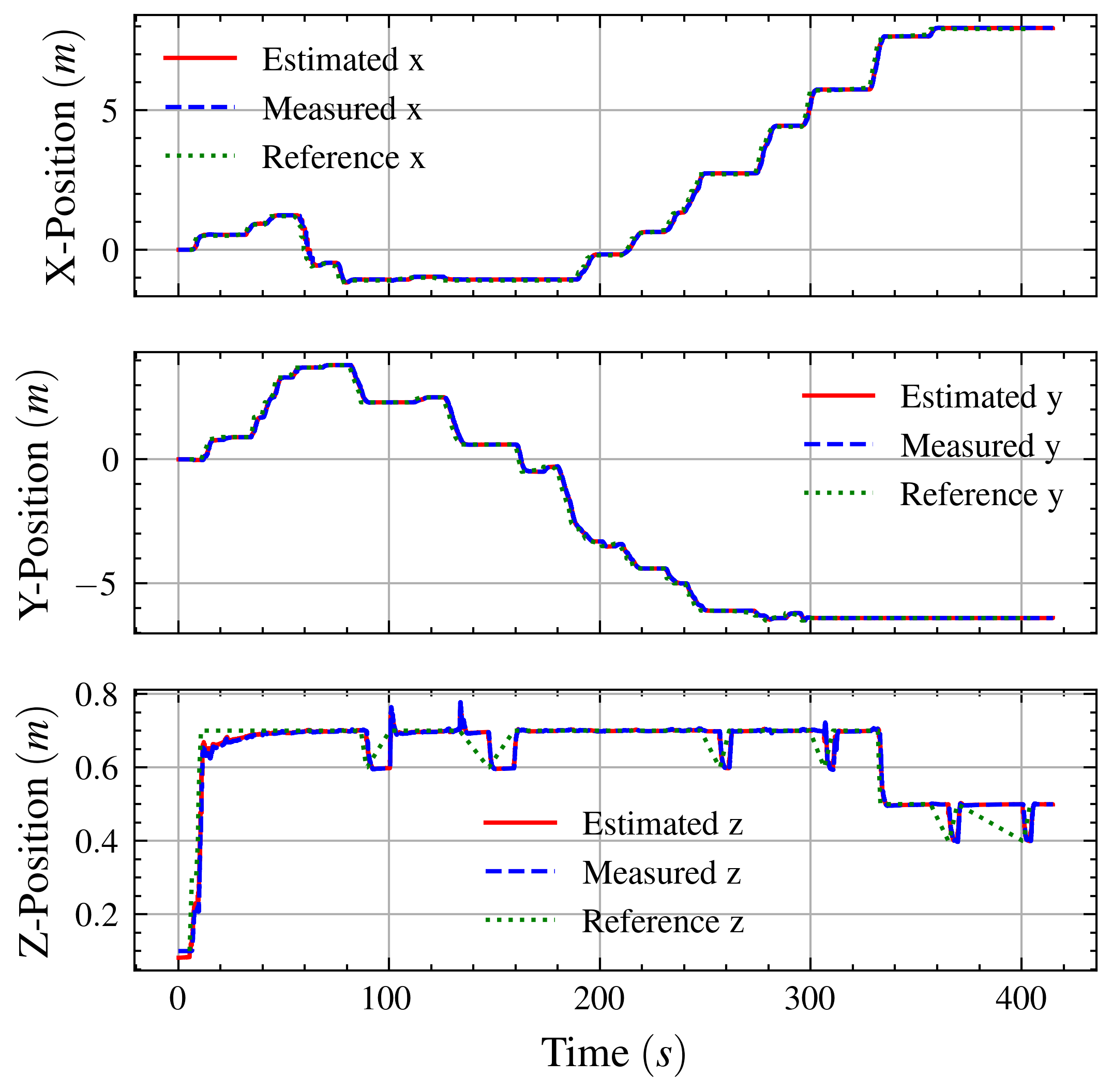}
     \caption{Estimated, measured, and reference position of the aerial robot in x, y, and z axis.}
     \label{fig:odom}
\end{figure}

\begin{figure}[http]
    \centering
    \includegraphics[width=\linewidth]{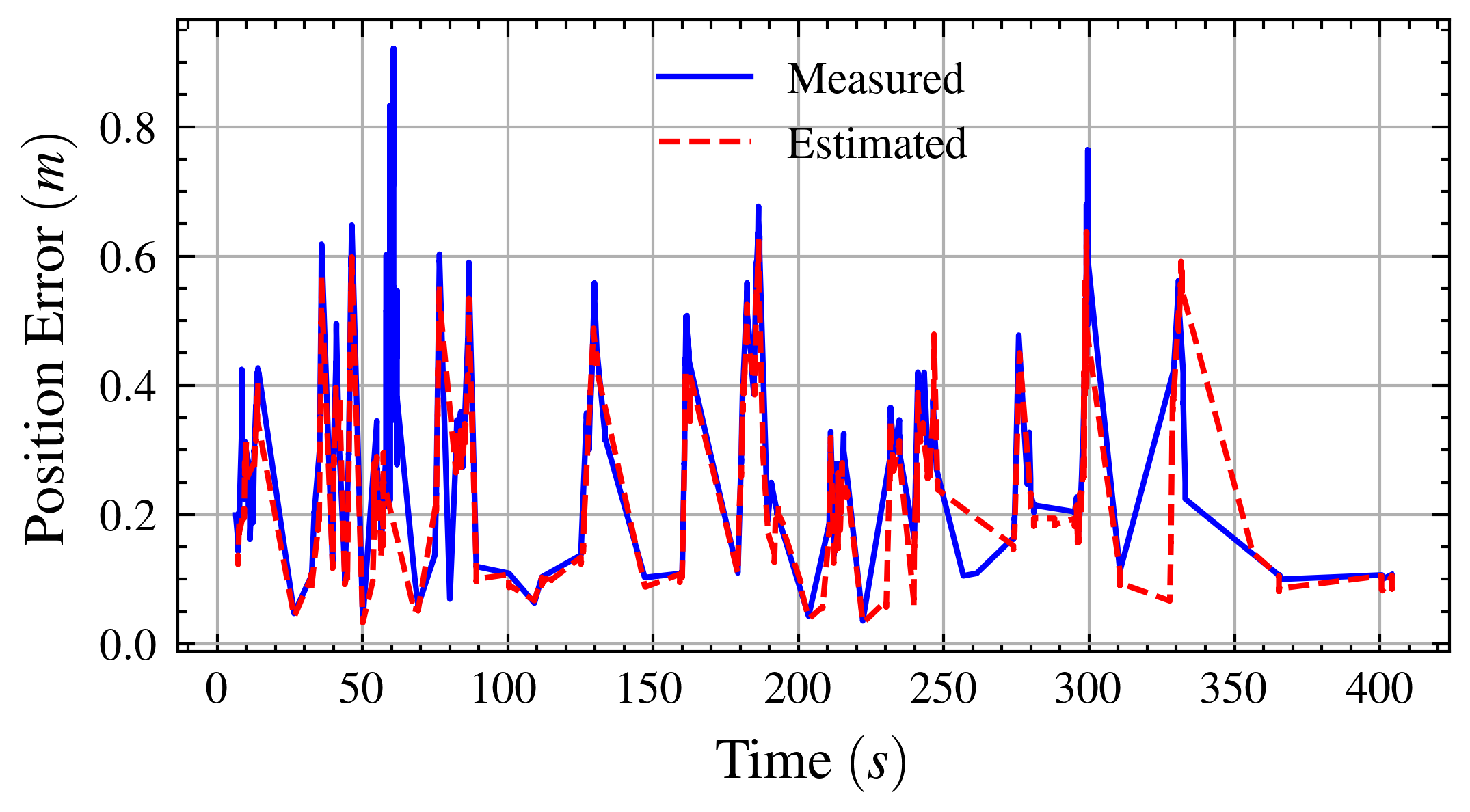}
     \caption{RMS values of position errors: reference vs. estimated and reference vs. measured.}
     \label{fig:error}
\end{figure}

Despite the network introduced delays, the cloud-assisted remote control successfully tracks the desired waypoints, as illustrated in Fig.~\ref{fig:odom}. The figure depicts the measured, estimated and reference values and showcases the ability of the proposed framework to follow the desired waypoints. The trajectory tracking performance is further evaluated through Root Mean Squared (RMS) error analysis of Fig.~\ref{fig:error}.

Fig.~\ref{fig:error} presents the RMS error values, demonstrating that the position predictor effectively estimates the actual robot position, resulting in lower reference-to-estimated errors compared to reference-to-measured errors. Notably, each time a new waypoint is introduced by the human operator, the error momentarily increases before the velocity controller compensates, maintaining errors below 0.1 cm.

\begin{figure*}[b]
    \centering
    \includegraphics[width=\linewidth]{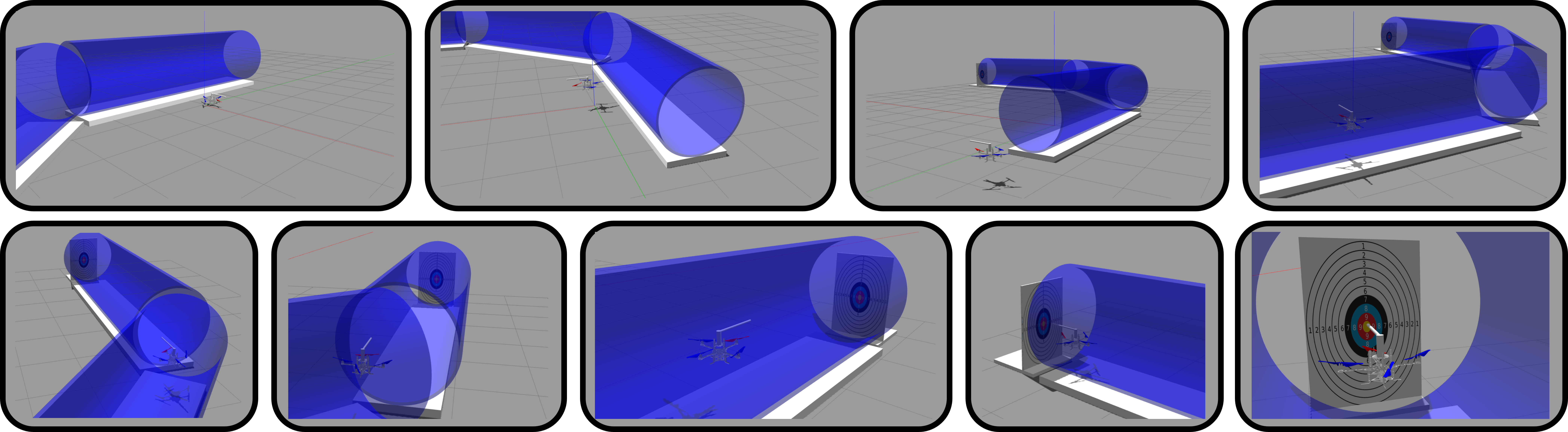}
     \caption{Snapshots of the Gazebo environment throughout the cloud-assisted remote control operation.}
     \label{fig:experiments}
\end{figure*}

Finally, snapshots of the Gazebo simulation during the cloud-assisted mission are presented in Fig.~\ref{fig:experiments}, illustrating key moments in the aerial robot’s trajectory. These visualizations further confirm the system's capability to operate effectively under cloud-based control, even in the presence of network-induced delays.

\section{Conclusions}
\label{sec:conclusions}
This paper introduces a cloud-emulated framework for aerial robotics, which enables cloud-assisted control via a simulated environment. The framework integrates containerized technologies, simulations using Gazebo, and network emulation to assess interactions between cloud services and robots under practically realistic scenarios. Utilizing a PID-based velocity controller, the system maintains precise trajectory tracking, even while compensating for network-induced delays with a position predictor. This emulated framework provides a controlled yet adaptable environment for testing and validating cloud robotics applications, thereby promoting research in cloud robotic systems.

\section{Acknowledgements}
\label{sec:acknowledgements}
We would like to thank the AERO-TRAIN project from the European Union’s Horizon 2020 research and innovation programme under the Marie Skłodowska-Curie grant agreement No 953454 for supporting this work.

\bibliographystyle{IEEEtran}
\bibliography{./IEEEtranBST/IEEEabrv, references}

\end{document}